\documentclass{article} 
\usepackage{iclr2027_conference,times}

\usepackage{amsmath,amsfonts,bm}

\def\eqref#1{equation~\ref{#1}}

\def\1{\bm{1}}

\DeclareMathAlphabet{\mathsfit}{\encodingdefault}{\sfdefault}{m}{sl}
\SetMathAlphabet{\mathsfit}{bold}{\encodingdefault}{\sfdefault}{bx}{n}

\usepackage{hyperref}
\usepackage{url}
\usepackage{booktabs}
\usepackage{graphicx}
\usepackage{xcolor}
\usepackage{listings}
\usepackage[most]{tcolorbox}

\title{Component-Aware Feedback for Self-Evolving Programs}

\author{
Ethan Lin \\
Santa Clara University \\
\texttt{enlin@scu.edu}
\And 
Jinming Nian \\
Santa Clara University \\
\texttt{jnian@scu.edu}
\And
Yi Fang \\
Santa Clara University \\
\texttt{yfang@scu.edu} 
}

\iclrfinalcopy 
\begin{document}

\maketitle
\lhead{Preprint. Under review.}

\begin{abstract}
LLM-guided evolutionary search can discover complex programs, but existing methods mostly only save candidate programs and fitness scores while discarding which component edits produced which fitness metric changes. Existing methods force the mutator LLM to infer the effect of prior edits from cluttered histories, making program search slow and unstable. This is especially true for locally servable LLMs to evolve multi-component systems. We introduce component-aware feedback, which compares each evaluated program with its parent, identifies the components that changed, and logs them with the associated metric differences into an attribution memory that later mutations read. The memory keeps each change in two reference frames, local against the parent it came from and global against the seed program, which shows both the immediate effect of a change and the cumulative progress made since the seed. We study this on LLM reranking, a multi-objective optimization problem where a multi-stage pipeline must balance quality against serving cost. Across twelve \textsc{Bright} datasets, our method reaches the strongest baseline's final quality after a median of one third of the search budget and ends 7.2\% higher in held-out nDCG@10, and under a cost-aware objective it finds pipelines that are on average more accurate while using 11\% fewer tokens per query, showing component-aware feedback to be a promising direction for more efficient self-evolving systems.

\end{abstract}

\section{Introduction}

Evolutionary search guided by large language models (LLMs) improves algorithms by proposing changes to executable programs, evaluating the resulting programs, and using the outcomes to guide further proposals. FunSearch and AlphaEvolve demonstrate the potential of this approach for mathematical and algorithmic discovery~\citep{funsearch,alphaevolve}. Since then, a growing line of systems has improved the loop in many ways: which parent to mutate, which population to grow, and to break out of local minima~\citep{shinka,cemri2026adaevolve}. The other half of the loop has received much less attention. Every iteration ends with an evaluation, and evaluations are the only signal the search learns from. As the evolved programs grow larger, one question becomes increasingly important: \textit{What information should each evaluation retain to best guide the mutator LLM in proposing the next change?} Today the answer is mostly a program and its score. The mutation prompt shows the parent, a few high scoring or diverse programs, their fitness, and the score changes of recent children~\citep{alphaevolve,shinka,cemri2026adaevolve}. For a program with several interacting components, the full pipeline code accompanied by an overall fitness score leaves substantial interpretation challenge to the mutator LLM. To tell which earlier edits helped, the mutator must diff full programs implicitly, decide which of the changed functions and constants mattered, and connect them to whichever metrics moved. It must do all of this from a context that grows longer and more cluttered as the search proceeds. LLMs are known to use long contexts unevenly and to lose track of accumulated multi-turn histories~\citep{liu2024lost,laban2026lost}. This burden matters most for the open, locally servable models that make evolutionary search affordable, whereas AlphaEvolve, ShinkaEvolve, and AdaEvolve all report their results with frontier models~\citep{alphaevolve,shinka,cemri2026adaevolve}. Reflection-based search and semantic-delta memories show that representing earlier modifications helps~\citep{reevo,deltaevolve}, but they ask an LLM to describe what an edit did, which is both expensive and hard to trust. The changes in component and the corresponding metric change can be programatically extracted. We argue that the mutator is best guided by these component-aware feedback.

We propose component-aware feedback for self-evolving programs, which turns each evaluation into a set of such records. After every evaluation, an \emph{observe} step decomposes the parent and the child into named components, namely their functions and module-level constants. It identifies which components were added, removed, or modified, and associates this edit set with the vector of differences in every metric the evaluator reports. The records accumulate in an \emph{attribution memory} that the mutation prompt reads, and each record is kept in two reference frames. When evolution is making steady progress, the \emph{local} frame compares a child with its parent and tells the mutator whether its last step helped. When search stagnates, the \emph{global} frame compares with seed and shows what has been attempted and changed so far. The records are built entirely from evaluations the search already performs, and we add them to AdaEvolve~\citep{cemri2026adaevolve} without changing its search controller. We study this approach on LLM reranking, which is an important problem on its own, and contains multiple components by nature. Reranking decides what a downstream model reads in search and retrieval-augmented systems. It remains an active research area, spanning prompting and log-likelihood methods~\citep{rankgpt,qin2024pairwise,chen2025tourrank} and rerankers trained with reinforcement learning~\citep{zhuang2026rankr1,liu2026reasonrank}, many of which are evaluated on the reasoning-intensive \textsc{Bright} benchmark~\citep{su2025bright}. LLM reranker is a pipeline whose components constrain one another. Retrieval sets the ceiling on achievable quality, and the reranking strategy decides how much of that ceiling is achieved, and at what computation cost. Evolving LLM reranking is often a trade off problem between quality and cost, where a good edit of one component may negatively impact the other. We therefore evolve the complete pipeline, and study a quality-only objective and a cost-aware one. 

\begin{figure}[t]
  \centering
  \includegraphics[width=\textwidth]{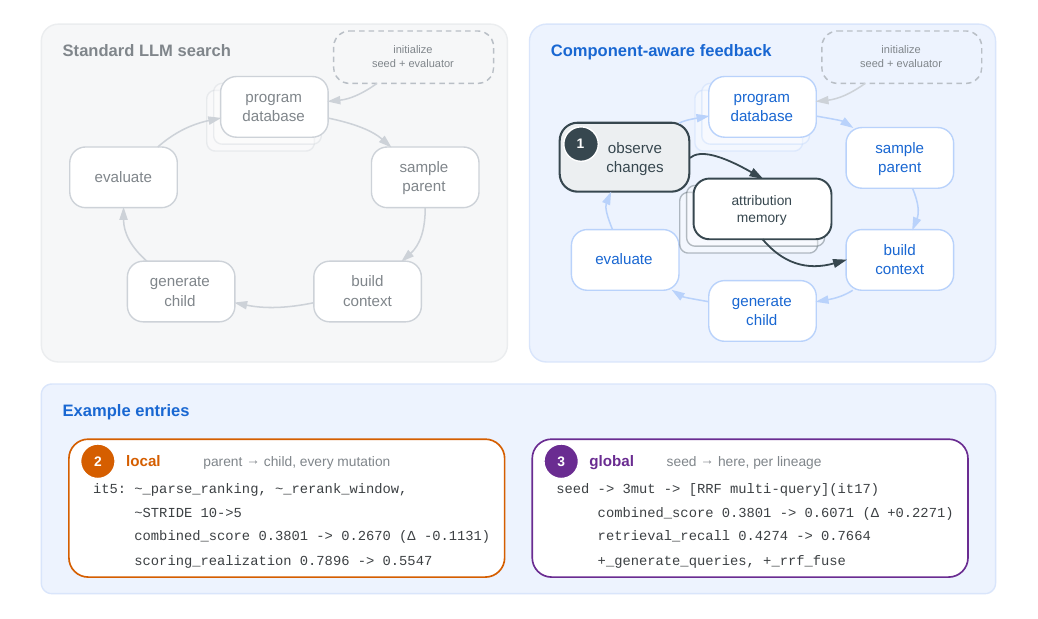}
  \caption{\textbf{Local and global component-aware feedback.} A standard LLM-guided search (top left) rebuilds its context from the database at every iteration and keeps nothing between iterations. We add an observe step (1), which diffs a child against its parent, and an attribution memory, which records each edit and is read by the context step (top right). Local and global entries (bottom) differ only in their reference point. A local entry (2) compares a child with its parent, and a global entry (3) compares it with the seed.}
  \label{fig:method}
\end{figure} 

We evaluate on all twelve \textsc{Bright} datasets, with one open-weight 35B model with 3B active parameters serving as both mutator and reranker~\citep{qwen36_35b_a3b}. With component-aware feedback, the search is faster, more consistent, and finds better programs. It raises held-out nDCG@10 by $7.2\%$ over AdaEvolve and leads on nine out of twelve datasets. On ten datasets, it reaches AdaEvolve's final search score after a median of 31 out of 100 iterations, and its final scores vary less across runs. Under the cost-aware objective, it finds pipelines that are more accurate and use $11\%$ fewer tokens per query compared to AdaEvolve. Its programs also transfer better. When run unchanged on other domains, they retain $90\%$ of the score obtained by evolving on the target directly, against $79\%$ for AdaEvolve. An ablation shows that the gains come from both attribution and from exposing more metrics. On circle packing and signal processing, our method is comparable to existing evolve methods, and works better on tasks where objectives compete with each other. These results suggest that when a program has many interacting components, telling the mutator which component changes moved which metrics is as important as deciding which programs to mutate.

\section{Methodology}
\subsection{Problem definition}
LLM-guided evolutionary search can be fundamentally broken down into a simple loop. A user initializes the search by providing a seed program $p_0$ and an evaluator with a fitness function $F$, and the system repeats five steps until a budget of $T$ iterations is reached:

\begin{enumerate}
    \item \textbf{Selection:} sample a parent program $p$ from the database $D$, together with a small set of inspiration programs.
    \item \textbf{Context:} assemble a prompt from the parent and inspiration programs.
    \item \textbf{Mutation:} generate a child program $p'$ from a mutation model $M$ given that prompt. 
    \item \textbf{Evaluation:} run the evaluator on the child program, obtaining a fitness score $f' = F(p')$.
    \item \textbf{Update:} add $p'$ into the database and update the controller that guides future selection.
\end{enumerate}

The goal of this loop is to maximize the fitness function $F : \mathcal{P} \rightarrow \mathbb{R}$ over the space of executable programs $\mathcal{P}$ by repeatedly generating and evaluating new programs $p_i$ over a fixed budget of $T$ iterations. Most systems maintain not one database but $K$ databases $D = \{D_1, D_2\dots, D_K\}$ that evolve semi-independently and periodically exchange programs, which keeps the design space from collapsing into a single lineage. Much of the effort in designing these systems goes into how to control the search to most efficiently utilize the fixed budget (e.g. which parent to sample, which database to spend the next iteration on, when to abandon a direction).

\subsection{Adaptive search control}
Prior works controlled the search with hand-designed parameters, fixing exploration rates, database counts and stagnation criteria ahead of a run and re-tuning them per problem. AdaEvolve~\citep{cemri2026adaevolve} removes this friction by deriving those parameters from the search itself. For each database $k$ it maintains an accumulated improvement signal, built from the normalized fitness gain $\delta^{(k)}_t$ that iteration $t$ produced,

\begin{equation}
    G^{(k)}_t \;=\; \rho\, G^{(k)}_{t-1} \;+\; (1-\rho)\big(\delta^{(k)}_t\big)^2
\label{eq:improvement_signal}
\end{equation}

with decay $\rho \in [0,1)$. Because each value is built from its predecessor, $G^{(k)}_t$ summarizes how productive that database has recently been. The search parameters then become functions of this signal rather than constants.

Despite these improvements, the content of the mutation prompt is not a function of any such signal. It is a fixed template of meta-instructions containing the full source code of the parent and a few selected programs as inspiration, the fitness scores of each, and a list of score deltas for the parent's earlier children. The improvement signal is used only to decide which parent is sampled, and the inspiration programs are selected by fitness scores and by how textually different they are from that parent. Everything in the template is therefore an information dump of source code and fitness scores, and none of it indicates the state the search is in the way the improvement signal does for the search parameters. That signal describes how the search is doing rather than what it has tried, so it cannot attribute which changes produced which effects. Thus in its current state, the mutation model is left to infer both the state of the search and the effect of every prior edit from a cluttered history. 

\subsection{Component-aware feedback}
Component-aware feedback removes this inference. For every program the search evaluates, it records which components the mutation changed and how each metric responded, and carries that record forward into the prompts that follow. The mutator can then see what has already been tried and what each attempt did, so that later mutations build on those outcomes instead of rediscovering them. We add an \emph{observe} step, which runs after each evaluation, and an \emph{attribution memory} that the context step reads (Figure~\ref{fig:method}). The context is then no longer an undifferentiated dump of programs and scores, but rather a record of which changes were made and what each one produced.

We decompose each program into named components, its functions and its module-level constants. A program is then a map from component names to source, where $p[n]$ is the source code of a component named $n$ and $\emptyset$ if $p$ has no component with that name. Observe diffs the parent's map against the child's to identify which components the mutation touched, and attributes the resulting change in every metric to that set,

\begin{equation}
\mathcal{C}_t \;=\; \big\{\, n \;\big|\; p[n] \neq p'[n] \,\big\},
\qquad
\Delta_t \;=\; \mu(p') - \mu(p),
\label{eq:edit}
\end{equation}

where $\mathcal{C}_t$ is the set of names $n$ for which the parent and child hold different source, and $\mu(\cdot)$ is the vector of metrics the evaluator returns. Three cases fall out of that single condition. A name in the child but not the parent is a component the mutation created, a name in the parent but not the child is one it removed, and a name in both with different source is one it modified. The pair $m_t = (\mathcal{C}_t, \Delta_t)$ is an \emph{edit}, and it is the unit the attribution memory stores. 

Edits accumulate in the attribution memory $\mathcal{A}$, which the context step reads alongside the database,

\begin{equation}
\text{prompt}_t \;=\; C\big(p,\, D_t,\, \mathcal{A}_t\big),
\qquad
\mathcal{A}_t \;=\; \mathcal{A}_{t-1} \cup \{m_t\},
\label{eq:memory}
\end{equation}

so the context depends on the whole trajectory of the search rather than only the programs that happened to be surfaced in $D_t$. The context becomes a function of the search state, as the search parameters already were in Eq.~\ref{eq:improvement_signal}. As with the improvement signal, one memory is kept per database and the two are merged whenever the databases exchange programs, so an edit made in one lineage becomes available to the other at migration rather than immediately. 

\subsection{Local and global reference frames}
An edit is only meaningful relative to a reference point. Applying Eq.~\ref{eq:edit} against the parent and against the seed gives two views that together show where the search stands,

\begin{equation}
m^{\,\text{loc}}_t \;=\; \big(\, \mathcal{C}^{\,\text{loc}}_t,\; \Delta^{\,\text{loc}}_t \,\big),
\qquad
m^{\,\text{glob}}_t \;=\; \big(\, \mathcal{C}^{\,\text{glob}}_t,\; \Delta^{\,\text{glob}}_t \,\big),
\label{eq:frames}
\end{equation}

where $\mathcal{C}^{\,\text{loc}}_t = \{\, n \mid p[n] \neq p'[n] \,\}$ and $\Delta^{\,\text{loc}}_t = \mu(p') - \mu(p)$ measure the child against its parent, and $\mathcal{C}^{\,\text{glob}}_t = \{\, n \mid p_0[n] \neq p'[n] \,\}$ and $\Delta^{\,\text{glob}}_t = \mu(p') - \mu(p_0)$ measure it against the seed.

The local frame measures an edit against its parent and answers whether that change helped. Local entries go into the mutation prompt as a sliding window of the ten most recent edits.

The global frame replaces the parent with the seed, so it describes a whole lineage rather than a single step. A program reached by a dozen mutations is compared directly against $p_0$, which says which components that entire chain has introduced or discarded and what the chain bought. Global entries are built only when the search stagnates, for the prompt that proposes a new strategy, where the shared origin makes those chains comparable and gives a direction time to pay off. 

The two frames can disagree, and an edit that lost ground against a strong parent is negative in $\Delta^{\,\text{loc}}$ while its lineage is still ahead in $\Delta^{\,\text{glob}}$, so the direction is worth continuing even though the last step was not. Figure~\ref{fig:method} shows an example entry in each frame. Appendix~\ref{appx:prompts} shows full prompts with each frame.

\subsection{Application to LLM reranking}
We study this on LLM reranking. A program $p$ maps a query $q$ to a ranking $\sigma$ over a corpus $\mathcal{D}$, retrieving its own candidates from $\mathcal{D}$ and then reordering them. The fitness score is ranking quality averaged over a query set $Q$,

\begin{equation}
p: (q, \mathcal{D}) \longmapsto \sigma,
\qquad
F(p) \;=\; \frac{1}{|Q|}\sum_{q \in Q} \text{nDCG}@10\big(p(q, \mathcal{D})\big).
\label{eq:rerank}
\end{equation}

Beyond nDCG@10 the evaluator reports the recall of the retrieved set, the nDCG an oracle reranker would achieve on it, the fraction of that ceiling the program achieves, and the cost of a query in LLM calls and tokens. Together these form the metric vector $\mu$ of Eq.~\ref{eq:edit}. 

A fitness score of nDCG@10 alone credits a program for any gain in ranking quality and never charges it for the LLM calls that produced the gain. We therefore also optimize a cost-aware objective that trades ranking quality against efficiency. We measure efficiency in LLM tokens per query. Every input or output token is one more the model must process or generate, so token usage is closely aligned with both the serving cost and the latency of a program. It measures neither directly, but it is a good proxy for how efficient the programs our search proposes are. We do not use wall-clock time because it depends on serving conditions such as resource contention from other processes, which vary over a run and would score the same program differently at different points in it. The objective combines a quality term $a$ and an efficiency term $b$, each normalized against the seed program,

\begin{equation}
s \;=\; 1 + \min(a, b) + \lambda\,(a+b),
\qquad
a = \frac{\text{nDCG} - \text{nDCG}_{0}}{1 - \text{nDCG}_{0}},
\qquad
b = 1 - \frac{\tau}{\tau_{0}},
\label{eq:cost}
\end{equation}

where $\tau$ is tokens per query, subscript $0$ denotes the seed, and $\lambda = 0.05$. The quality term $a$ is the gain in nDCG@10 over the seed as a fraction of the largest gain possible from it, and the efficiency term $b$ is the fraction of the seed's tokens per query the program saves. Both are zero at the seed and one at their ceiling, and the constant places the seed at a score of 1. We clip each term to $[-1,1]$, since a program that uses far more tokens than the seed would otherwise have an unbounded efficiency term. The min term directs the search toward whichever of quality and efficiency is weaker. Without the $\lambda$ term, the search could spend its whole budget fixated on either quality or efficiency, when the goal is the improve both. The $\lambda$ term adds a small bonus for gains in both, so progress on either objective is rewarded.

\section{Experiments}
\subsection{Experimental Setup} \label{sec:setup}
We evaluate on \textsc{Bright}~\citep{su2025bright}, twelve retrieval datasets spanning StackExchange domains, coding, and theorem-based questions. Each dataset is split 60/40 into train and test. The search optimizes a fitness score evaluated on the train split, either nDCG@10 or the cost-aware score of Eq.~\ref{eq:cost}. We report held-out test nDCG@10 of the final program, mean, and standard deviation over three runs per dataset. 

The seed program is BM25~\citep{bm25} over the corpus followed by a sliding-window listwise reranker~\citep{rankgpt}. All methods share this seed, evaluator, budget of $T=100$ iterations, model for both mutation and reranking, $K=2$ databases with migration every 10 iterations, and four context programs per prompt. The mutator samples at temperature $0.7$ with up to $16{,}384$ output tokens, and the reranking calls inside evolved programs decode greedily. The main results use Qwen3.6-35B-A3B~\citep{qwen36_35b_a3b}. We compare against BM25, the seed program, and AdaEvolve~\citep{cemri2026adaevolve}. Appendix~\ref{appx:programs} shows the seed and examples of evolved programs under both objectives.

\subsection{Experimental Results} 
\textbf{Ranking quality and cost.} \label{sec:cost} Table~\ref{tab:bright-main} reports held-out nDCG@10 under the quality-only objective. With component-aware feedback, the search returns programs that average $0.346$ against AdaEvolve's $0.323$, a relative gain of $7.2\%$, and leads on nine of the twelve datasets. Both searches far exceed the seed at $0.257$ and BM25 at $0.140$.

Under the quality-only objective this improvement comes at a cost. These programs use $71.8$k LLM tokens per query averaged over the twelve datasets against AdaEvolve's $53.7$k. A component that widens the candidate pool is credited with higher ranking quality and is never penalized for adding more LLM calls or tokens per query. Thus, an efficient pipeline and an expensive one are indistinguishable under a quality-only objective. Component-aware feedback finds the better-ranking program at the expense of cost and latency because nothing in the objective promotes efficiency. Exploring the trade-off between ranking quality and efficiency therefore requires a fitness score that accounts for both.

\begin{table*}[t]
\centering
\small
\setlength{\tabcolsep}{4pt}
\caption{\textbf{Ranking performance on \textsc{Bright}.} nDCG@10 on the train and held-out test splits, split 60/40 per dataset. Best is the best single run on test. $\Delta$\% is our relative gain over AdaEvolve on test. Mean $\pm$ std over 3 runs.}
\label{tab:bright-main}
\resizebox{\textwidth}{!}{%
\begin{tabular}{@{}l cc cc ccc ccc c@{}}
\toprule
& \multicolumn{2}{c}{BM25} & \multicolumn{2}{c}{seed program}
& \multicolumn{3}{c}{AdaEvolve} & \multicolumn{3}{c}{\textbf{Ours}} & \\
\cmidrule(lr){2-3}\cmidrule(lr){4-5}\cmidrule(lr){6-8}\cmidrule(lr){9-11}
dataset & train & test & train & test & train & test & best & train & test & best & $\Delta$\% \\
\midrule
\multicolumn{12}{@{}l}{\textit{StackExchange}}\\
bio.    & 0.184 & 0.167 & $0.372${\scriptsize$\pm$.006} & 0.362 & $0.572${\scriptsize$\pm$.027} & $0.581${\scriptsize$\pm$.044} & 0.619 & $\mathbf{0.602}${\scriptsize$\pm$.005} & $\mathbf{0.613}${\scriptsize$\pm$.021} & \textbf{0.633} & $+5.6$ \\
earth.  & 0.238 & 0.363 & $0.404${\scriptsize$\pm$.012} & 0.509 & $0.492${\scriptsize$\pm$.016} & $0.559${\scriptsize$\pm$.014} & 0.575 & $\mathbf{0.529}${\scriptsize$\pm$.020} & $\mathbf{0.573}${\scriptsize$\pm$.035} & \textbf{0.606} & $+2.4$ \\
econ.   & 0.180 & 0.118 & $0.299${\scriptsize$\pm$.009} & 0.201 & $0.407${\scriptsize$\pm$.024} & $0.253${\scriptsize$\pm$.030} & 0.278 & $\mathbf{0.412}${\scriptsize$\pm$.018} & $\mathbf{0.274}${\scriptsize$\pm$.023} & \textbf{0.299} & $+8.1$ \\
psy.    & 0.089 & 0.228 & $0.307${\scriptsize$\pm$.006} & 0.501 & $0.434${\scriptsize$\pm$.023} & $\mathbf{0.540}${\scriptsize$\pm$.035} & 0.563 & $\mathbf{0.448}${\scriptsize$\pm$.008} & $0.533${\scriptsize$\pm$.066} & \textbf{0.604} & $-1.3$ \\
rob.    & 0.154 & 0.075 & $0.337${\scriptsize$\pm$.006} & 0.239 & $0.376${\scriptsize$\pm$.014} & $0.254${\scriptsize$\pm$.015} & 0.268 & $\mathbf{0.404}${\scriptsize$\pm$.025} & $\mathbf{0.275}${\scriptsize$\pm$.005} & \textbf{0.281} & $+8.4$ \\
stack.  & 0.199 & 0.142 & $0.322${\scriptsize$\pm$.016} & 0.219 & $0.406${\scriptsize$\pm$.029} & $0.211${\scriptsize$\pm$.044} & 0.252 & $\mathbf{0.424}${\scriptsize$\pm$.011} & $\mathbf{0.219}${\scriptsize$\pm$.040} & \textbf{0.260} & $+3.9$ \\
sus.    & 0.172 & 0.137 & $0.267${\scriptsize$\pm$.009} & 0.300 & $0.314${\scriptsize$\pm$.029} & $0.302${\scriptsize$\pm$.060} & 0.363 & $\mathbf{0.353}${\scriptsize$\pm$.026} & $\mathbf{0.321}${\scriptsize$\pm$.053} & \textbf{0.377} & $+6.3$ \\
\midrule
\multicolumn{12}{@{}l}{\textit{Coding}}\\
leet.   & 0.303 & 0.196 & $0.289${\scriptsize$\pm$.004} & 0.190 & $0.360${\scriptsize$\pm$.016} & $0.207${\scriptsize$\pm$.014} & 0.220 & $\mathbf{0.381}${\scriptsize$\pm$.031} & $\mathbf{0.247}${\scriptsize$\pm$.017} & \textbf{0.264} & $+19.1$ \\
pony    & 0.052 & 0.047 & $0.165${\scriptsize$\pm$.011} & 0.150 & $\mathbf{0.267}${\scriptsize$\pm$.011} & $\mathbf{0.220}${\scriptsize$\pm$.032} & \textbf{0.254} & $0.248${\scriptsize$\pm$.003} & $0.204${\scriptsize$\pm$.001} & 0.205 & $-7.0$ \\
\midrule
\multicolumn{12}{@{}l}{\textit{Theorem-based}}\\
aops    & 0.037 & 0.098 & $0.057${\scriptsize$\pm$.004} & 0.130 & $\mathbf{0.081}${\scriptsize$\pm$.013} & $\mathbf{0.119}${\scriptsize$\pm$.028} & \textbf{0.145} & $0.080${\scriptsize$\pm$.019} & $0.087${\scriptsize$\pm$.025} & 0.115 & $-26.4$ \\
theoQ.  & 0.086 & 0.071 & $0.129${\scriptsize$\pm$.006} & 0.124 & $0.314${\scriptsize$\pm$.060} & $0.286${\scriptsize$\pm$.048} & 0.325 & $\mathbf{0.381}${\scriptsize$\pm$.037} & $\mathbf{0.348}${\scriptsize$\pm$.054} & \textbf{0.380} & $+21.7$ \\
theoT.  & 0.004 & 0.037 & $0.063${\scriptsize$\pm$.002} & 0.157 & $0.421${\scriptsize$\pm$.131} & $0.344${\scriptsize$\pm$.130} & 0.494 & $\mathbf{0.536}${\scriptsize$\pm$.022} & $\mathbf{0.459}${\scriptsize$\pm$.033} & \textbf{0.496} & $+33.5$ \\
\midrule
mean    & 0.141 & 0.140 & 0.251 & 0.257 & 0.370 & 0.323 & 0.363 & \textbf{0.400} & \textbf{0.346} & \textbf{0.377} & $+7.2$ \\
\bottomrule
\end{tabular}
}
\vspace{-10pt}
\end{table*}

\begin{table*}[t]
\centering
\small
\setlength{\tabcolsep}{4pt}
\caption{\textbf{Cost-aware search on \textsc{Bright}.} Test nDCG@10 and LLM tokens per query of the final program. Both searches optimize the cost-aware objective rather than nDCG alone. $\Delta$ is the change in test nDCG@10 against Table~\ref{tab:bright-main}. Cheaper is the quality-only program's tokens per query divided by this program's. Mean $\pm$ std over 3 runs.}
\label{tab:bright-cost}
\begin{tabular}{@{}l cccc cccc@{}}
\toprule
& \multicolumn{4}{c}{AdaEvolve} & \multicolumn{4}{c}{\textbf{Ours}} \\
\cmidrule(lr){2-5}\cmidrule(lr){6-9}
dataset & nDCG@10 & $\Delta$ & tok/q & cheaper & nDCG@10 & $\Delta$ & tok/q & cheaper \\
\midrule
\multicolumn{9}{@{}l}{\textit{StackExchange}}\\
bio.    & $0.542${\scriptsize$\pm$.012} & -0.039 & \textbf{10,118} & 6.2x & $\mathbf{0.558}${\scriptsize$\pm$.010} & -0.055 & 12,071 & 15.1x \\
earth.  & $0.521${\scriptsize$\pm$.021} & -0.038 & 27,539 & 2.2x & $\mathbf{0.540}${\scriptsize$\pm$.030} & -0.033 & \textbf{14,930} & 2.6x \\
econ.   & $\mathbf{0.225}${\scriptsize$\pm$.007} & -0.029 & 14,453 & 5.2x & $0.217${\scriptsize$\pm$.044} & -0.056 & \textbf{7,055} & 5.2x \\
psy.    & $0.441${\scriptsize$\pm$.031} & -0.099 & \textbf{10,422} & 5.6x & $\mathbf{0.488}${\scriptsize$\pm$.019} & -0.045 & 25,736 & 2.8x \\
rob.    & $\mathbf{0.238}${\scriptsize$\pm$.010} & -0.016 & 45,186 & 1.6x & $0.226${\scriptsize$\pm$.038} & -0.049 & \textbf{23,025} & 5.4x \\
stack.  & $\mathbf{0.202}${\scriptsize$\pm$.057} & -0.009 & 16,518 & 3.8x & $0.185${\scriptsize$\pm$.077} & -0.035 & \textbf{10,693} & 3.4x \\
sus.    & $0.297${\scriptsize$\pm$.048} & -0.004 & 14,505 & 1.2x & $\mathbf{0.341}${\scriptsize$\pm$.027} & +0.020 & \textbf{12,466} & 2.6x \\
\midrule
\multicolumn{9}{@{}l}{\textit{Coding}}\\
leet.   & $\mathbf{0.232}${\scriptsize$\pm$.014} & +0.024 & 7,553 & 3.3x & $0.219${\scriptsize$\pm$.029} & -0.029 & \textbf{6,920} & 5.2x \\
pony    & $\mathbf{0.166}${\scriptsize$\pm$.113} & -0.054 & \textbf{7,635} & 8.3x & $0.141${\scriptsize$\pm$.013} & -0.063 & 12,548 & 7.4x \\
\midrule
\multicolumn{9}{@{}l}{\textit{Theorem-based}}\\
aops    & $0.090${\scriptsize$\pm$.048} & -0.029 & \textbf{2,151} & 37.5x & $\mathbf{0.091}${\scriptsize$\pm$.019} & +0.003 & 8,847 & 3.7x \\
theoQ.  & $0.218${\scriptsize$\pm$.020} & -0.069 & 6,398 & 4.7x & $\mathbf{0.264}${\scriptsize$\pm$.007} & -0.084 & \textbf{5,049} & 12.6x \\
theoT.  & $0.379${\scriptsize$\pm$.067} & +0.036 & \textbf{8,432} & 3.9x & $\mathbf{0.393}${\scriptsize$\pm$.064} & -0.066 & 12,765 & 9.0x \\
\midrule
mean    & 0.296 & -0.027 & 14,243 & 7.0x & 0.305 & -0.041 & 12,676 & 6.2x \\
\bottomrule
\end{tabular}
\vspace{-10pt}
\end{table*}

The cost-aware objective of Eq.~\ref{eq:cost} scores ranking quality and token usage together. Under it, both searches return programs six to seven times cheaper than under quality alone (Table~\ref{tab:bright-cost}). Component-aware feedback leads on both terms, reaching $0.305$ nDCG@10 at $12{,}676$ tokens per query against AdaEvolve's $0.296$ at $14{,}243$, an $11\%$ saving at a higher score.

\begin{figure}[t]
  \centering
  \includegraphics[width=\textwidth]{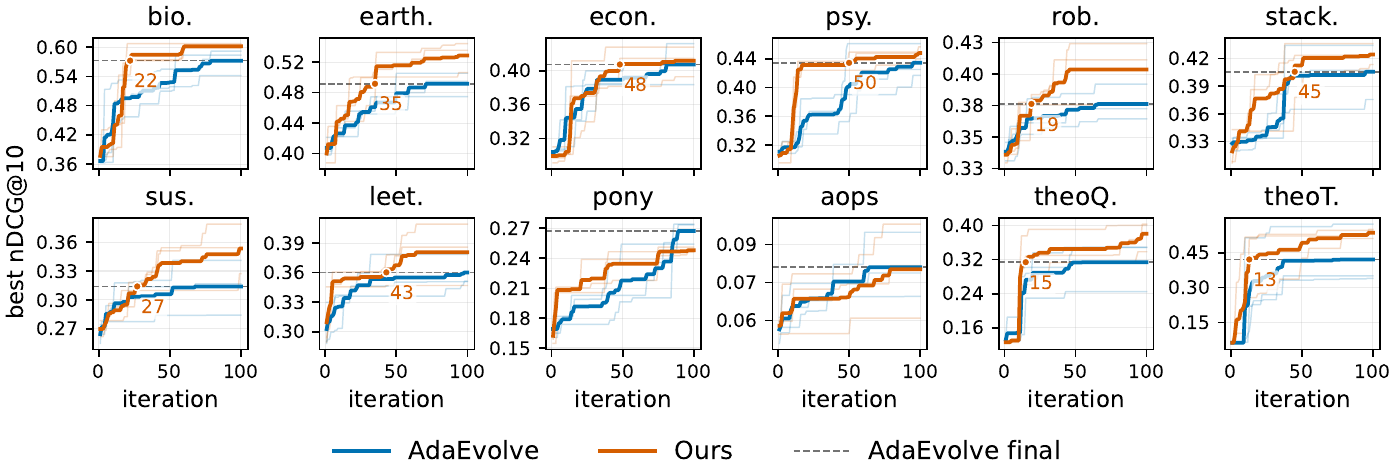}
    \caption{\textbf{Convergence on \textsc{Bright}.} Best-so-far nDCG@10 on the train split, one panel per dataset. Solid lines are the mean over 3 runs, faint lines the individual runs. The marker gives the iteration at which our mean first reaches AdaEvolve's final score.}
  \label{fig:convergence}
  \vspace{-10pt}
\end{figure}

\begin{figure}[t]
  \centering
  \includegraphics[width=\textwidth]{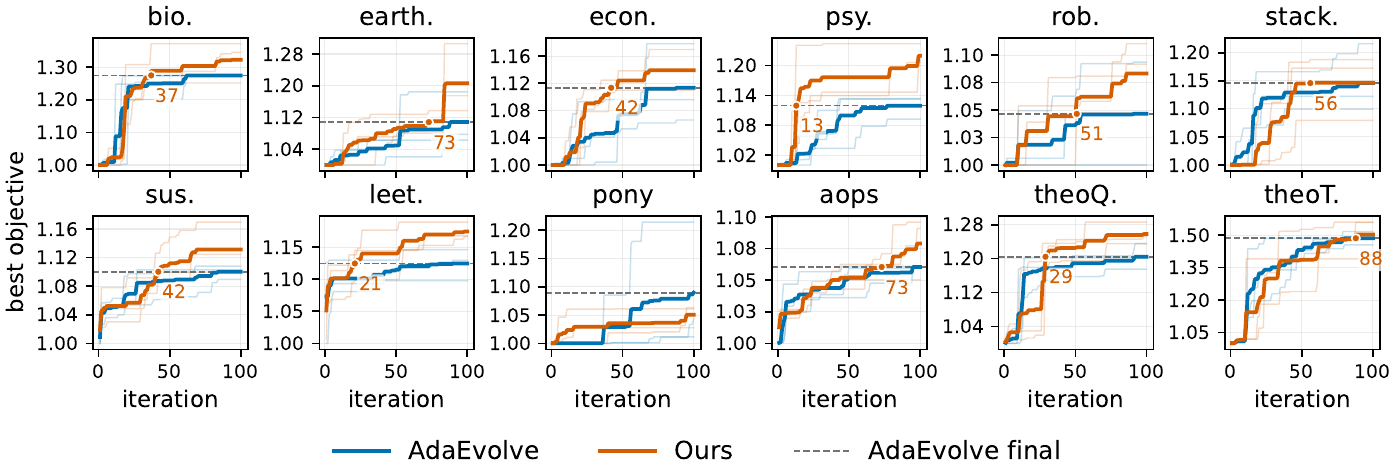}
    \caption{\textbf{Convergence under the cost-aware objective.} Best-so-far objective value on the train split, one panel per dataset. The seed program scores 1. Solid lines are the mean over 3 runs, faint lines the individual runs. The marker gives the iteration at which our mean first reaches AdaEvolve's final value.}
  \label{fig:convergence-cost}
  \vspace{-10pt}
\end{figure}

\textbf{Convergence speed.} \label{sec:convergence} If the mutator no longer has to infer the effect of each prior edit from a cluttered history, the search should reach a given score in fewer iterations. Figures~\ref{fig:convergence} and~\ref{fig:convergence-cost} plot the best-so-far train score against iteration, one panel per dataset. Under the quality-only objective our mean reaches AdaEvolve's final score on ten of the twelve datasets, at a median of $31$ iterations out of $100$ and a mean of $32$. Under the cost-aware objective it does so on eleven of twelve, at a median of $42$ and a mean of $48$. Our method's curve also stays above AdaEvolve's for most of the run on eleven of the twelve datasets under both objectives.

Both searches plateau well before the budget runs out, improving by a median of under $1\%$ over the final quarter under the quality-only objective, so most of what a search will find it finds early. Each iteration pairs a mutation call with an evaluation, and in most applications of evolutionary search the evaluation dominates the cost, whether it runs a pipeline over a benchmark, trains a model, or executes a simulation. A search that reaches a given score in a third of the iterations therefore does so for roughly a third of the compute, and under a fixed compute budget spends more of that budget searching from better programs.

\textbf{Cross-domain generalization.} \label{sec:transfer} To test whether the programs component-aware feedback finds are specific to the dataset they were evolved on, we take the three programs each method optimized on biology under the quality-only objective and run them unchanged on four other datasets. The four datasets differ in how closely they resemble biology. Earth science is a neighboring natural science, robotics shares biology's StackExchange source, and leetcode and theoremqa come from the coding and theorem-based sections of the \textsc{Bright} benchmark. Each is scored on the held-out split. 

Component-aware feedback finds programs that generalize better across datasets. Both methods lose ranking quality when we applied their biology programs to a new dataset, but ours lose less on every dataset (Table~\ref{tab:bright-transfer}) and keep $90\%$ of their in-domain score on average compared to $79\%$ for AdaEvolve, even matching the in-domain programs on robotics.

\begin{table*}[t]
\centering
\small
\setlength{\tabcolsep}{6pt}
\caption{\textbf{Cross-domain generalization on \textsc{Bright}.} Programs optimized on biology scored on other datasets' held-out splits (nDCG@10). Seed and optimized on that dataset are the unevolved program and each method's search run on that dataset respectively, all from Table~\ref{tab:bright-main}. Retained is the transferred program's score as a percentage of the same method's score when optimized on that dataset. Mean $\pm$ std over 3 runs.}
\label{tab:bright-transfer}
\begin{tabular}{@{}l c cc cc cc@{}}
\toprule
 & & \multicolumn{2}{c}{optimized on that dataset} & \multicolumn{2}{c}{optimized on biology} & \multicolumn{2}{c}{retained} \\
\cmidrule(lr){3-4}\cmidrule(lr){5-6}\cmidrule(lr){7-8}
dataset & seed & AdaEvolve & \textbf{Ours} & AdaEvolve & \textbf{Ours} & AdaEvolve & \textbf{Ours} \\
\midrule
earth. & $0.509$ & $0.559${\scriptsize$\pm$.014} & $\mathbf{0.573}${\scriptsize$\pm$.035} & $0.517${\scriptsize$\pm$.029} & $\mathbf{0.548}${\scriptsize$\pm$.037} & $92\%$ & $\mathbf{96\%}$ \\
rob. & $0.239$ & $0.254${\scriptsize$\pm$.015} & $\mathbf{0.275}${\scriptsize$\pm$.005} & $0.220${\scriptsize$\pm$.024} & $\mathbf{0.284}${\scriptsize$\pm$.028} & $87\%$ & $\mathbf{103\%}$ \\
leet. & $0.190$ & $0.207${\scriptsize$\pm$.014} & $\mathbf{0.247}${\scriptsize$\pm$.017} & $0.126${\scriptsize$\pm$.056} & $\mathbf{0.177}${\scriptsize$\pm$.022} & $61\%$ & $\mathbf{71\%}$ \\
theoQ. & $0.124$ & $0.286${\scriptsize$\pm$.048} & $\mathbf{0.348}${\scriptsize$\pm$.054} & $0.222${\scriptsize$\pm$.104} & $\mathbf{0.311}${\scriptsize$\pm$.108} & $78\%$ & $\mathbf{89\%}$ \\
\midrule
mean & $0.266$ & $0.327$ & $\mathbf{0.361}$ & $0.271$ & $\mathbf{0.330}$ & $79\%$ & $\mathbf{90\%}$ \\
\bottomrule
\end{tabular}
\vspace{-10pt}
\end{table*}

\begin{figure}[t]
  \centering
  \includegraphics[width=\textwidth]{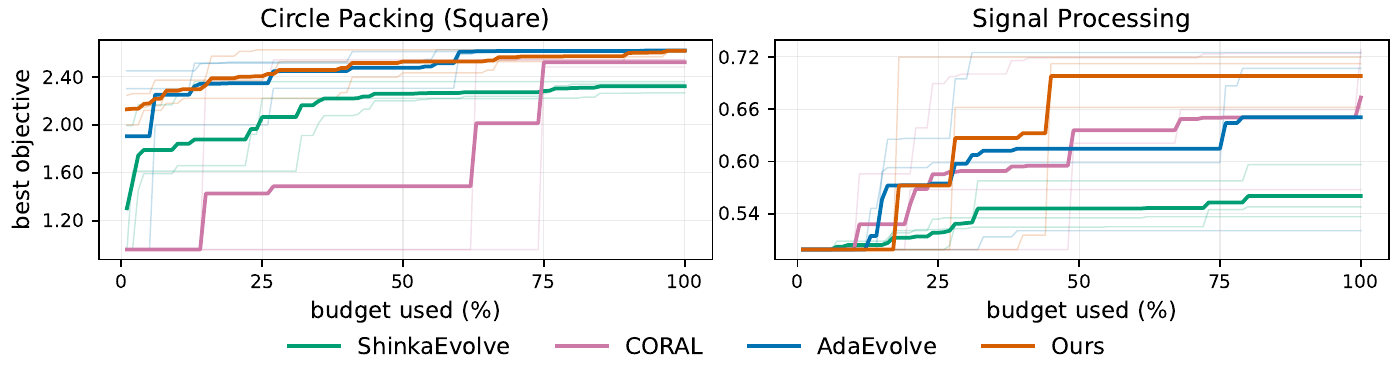}
  \caption{\textbf{Mathematical optimization.} Best-so-far objective against the fraction of the search budget spent, one panel per benchmark, higher is better. Solid lines are the mean over 3 runs, faint lines the individual runs.}
  \label{fig:math}
  \vspace{-10pt}
\end{figure}

\textbf{Beyond retrieval.} \label{sec:math} Beyond retrieval, we evaluate component-aware feedback on two mathematical optimization benchmarks (Figure~\ref{fig:math}). Circle packing places $26$ circles in a unit square to maximize the sum of their radii. Signal processing evolves a filter that removes noise from a changing signal while tracking it closely, scored on a combination of several measures of smoothness, accuracy and noise removal. Alongside AdaEvolve we compare against ShinkaEvolve~\citep{shinka}, another LLM-driven evolutionary search, and CORAL~\citep{coral}, a multi-agent system of autonomous coding agents. CORAL has no notion of an iteration, so we give it a budget equal to the average cost of one of our runs.

On circle packing, component-aware feedback matches AdaEvolve, $2.617$ against $2.619$, with the same small spread acorss runs, and both lead CORAL at $2.522$ and ShinkaEvolve at $2.322$. Circle packing has a single score, so attribution adds little to the score change AdaEvolve already reports. On signal processing, component-aware feedback is more consistent than AdaEvolve and CORAL. Its three runs stay between $0.662$ and $0.720$, while AdaEvolve and CORAL each have a run that drops to $0.521$ and $0.568$, which gives component-aware feedback the highest mean, $0.698$ versus $0.673$ for CORAL, $0.651$ for AdaEvolve and $0.560$ for ShinkaEvolve. Unlike circle packing, signal processing has competing goals, since a filter that removes more noise also responds more slowly to real changes. Attribution shows the mutator which edits moved which measure, which is what balancing these competing objectives requires, as with quality and cost in reranking. 

\textbf{Ablations. } \label{sec:ablation} Component-aware feedback has two parts, the evaluator's decomposed metrics and their attribution to the components that changed. We isolate each part to test its value on biology, leetcode, and theoremqa, one dataset from each section of the \textsc{Bright} benchmark. We compare against AdaEvolve and only evolving the reranking prompt. Evolving the whole program raises test nDCG@10 from $0.245$ to $0.358$. Decomposed metrics raise it to $0.382$ and attribution to $0.403$, so each part accounts for about half of our gain over AdaEvolve. Attribution adds little on train, $0.454$ versus $0.448$, but narrows the drop from train to test from $0.066$ to $0.051$.

\begin{table*}[t]
\centering
\small
\setlength{\tabcolsep}{4pt}
\caption{\textbf{Component ablation on \textsc{Bright}.} nDCG@10 on one dataset per \textsc{Bright} section. Prompt evolution evolves the reranking prompt, and the other rows evolve the program. Mean $\pm$ std over 3 runs.}
\label{tab:bright-ablation}
\resizebox{\textwidth}{!}{%
\begin{tabular}{@{}l cc cc cc cc@{}}
\toprule
 & \multicolumn{2}{c}{bio.} & \multicolumn{2}{c}{leet.} & \multicolumn{2}{c}{theoQ.} & \multicolumn{2}{c}{mean} \\
\cmidrule(lr){2-3}\cmidrule(lr){4-5}\cmidrule(lr){6-7}\cmidrule(lr){8-9}
method & train & test & train & test & train & test & train & test \\
\midrule
prompt evolution & $0.408${\scriptsize$\pm$.012} & $0.373${\scriptsize$\pm$.011} & $0.376${\scriptsize$\pm$.007} & $0.221${\scriptsize$\pm$.019} & $0.146${\scriptsize$\pm$.001} & $0.140${\scriptsize$\pm$.004} & $0.310$ & $0.245$ \\
AdaEvolve & $0.572${\scriptsize$\pm$.027} & $0.581${\scriptsize$\pm$.044} & $0.360${\scriptsize$\pm$.016} & $0.207${\scriptsize$\pm$.014} & $0.314${\scriptsize$\pm$.060} & $0.286${\scriptsize$\pm$.048} & $0.416$ & $0.358$ \\
\textbf{Ours} & $0.602${\scriptsize$\pm$.005} & $0.613${\scriptsize$\pm$.021} & $0.381${\scriptsize$\pm$.031} & $\mathbf{0.247}${\scriptsize$\pm$.017} & $\mathbf{0.381}${\scriptsize$\pm$.037} & $\mathbf{0.348}${\scriptsize$\pm$.054} & $\mathbf{0.454}$ & $\mathbf{0.403}$ \\
\quad $-$ attribution & $\mathbf{0.609}${\scriptsize$\pm$.036} & $\mathbf{0.635}${\scriptsize$\pm$.060} & $\mathbf{0.392}${\scriptsize$\pm$.023} & $0.208${\scriptsize$\pm$.004} & $0.343${\scriptsize$\pm$.073} & $0.302${\scriptsize$\pm$.030} & $0.448$ & $0.382$ \\
\bottomrule
\end{tabular}
}
\vspace{-10pt}
\end{table*}

\section{Related Work}
\textbf{LLM-guided program evolution.} FunSearch and AlphaEvolve established the basic loop of LLM mutation, program evaluation, and a program database~\citep{funsearch,alphaevolve}. EoH and ReEvo extended the loop to heuristic design~\citep{eoh2024,reevo}, and OpenEvolve provides an open implementation of AlphaEvolve~\citep{openevolve}. Most later work improves search control. ShinkaEvolve does so through parent sampling, novelty rejection, and model selection~\citep{shinka}, AdaEvolve through adaptive exploration and budget allocation~\citep{cemri2026adaevolve}, and EvoX by evolving the search strategy itself~\citep{evox2026}. CORAL and SwarmResearch instead hand the loop to autonomous coding agents~\citep{coral,swarmresearch2026}. Our work is orthogonal to search control as we focus on the feedback provided to the mutator. 

\textbf{Feedback for LLM optimizers}
OpenEvolve and AdaEvolve shows past solutions and their scores in the prompt. ReEvo and ShinkaEvolve summarize past attempts in LLM-written reflections~\citep{reevo,shinka}. For compound AI systems, MIPRO, TextGrad, Trace, and GEPA assign credit to individual modules using surrogate models, textual gradients, execution traces, and reflection~\citep{opsahlong2024mipro,yuksekgonul2025textgrad,cheng2024trace,gepa2026}. DeltaEvolve describes each parent to child change in text and pairs it with a qualitative signal of improvement~\citep{deltaevolve}. Component-aware feedback instead reads directly from the source code diff and the evaluator, so we do not need LLM summaries, extra evaluations, or learned models. 

\textbf{LLM reranking.} Zero-shot LLM rerankers differ in how they elicit relevance. Pointwise methods use the likelihood of relevance labels or of the query~\citep{nogueira2020monot5,zhuang2024beyond,sachan2022upr}, or attention patterns~\citep{chen2025icr}. Comparative methods use pairwise, setwise, listwise, or tournament prompting~\citep{qin2024pairwise,zhuang2024setwise,rankgpt,reddy2024first,chen2025tourrank}. Another line of work trains rerankers, through distillation~\citep{pradeep2023rankzephyr,weller2025rank1} or reinforcement learning~\citep{zhuang2026rankr1,zhang2025rearank,liu2026reasonrank,cai2026erank}, and these rerankers are often evaluated on \textsc{Bright}~\citep{su2025bright}. JudgeRank and DIVER combine several reasoning or retrieval stages into one pipeline~\citep{niu2024judgerank,sun2025diver}. Each of these methods is a hand-designed point in a space of retrieval and reranking choices. We instead search that space with a fixed, locally served LLM. Because all methods we compare share one seed, one model, and one evaluation protocol, our results measure the search itself and are not meant to compete with published \textsc{Bright} numbers.

\section{Conclusion}
Most work on LLM-guided evolutionary search refines how the search is controlled. We asked instead what each evaluation should retain to guide the mutator. Our answer is component-aware feedback, a record of which components each edit changed and their associated outcomes. With it, a locally served model reaches the strongest baseline's final score in a third of the iterations and evolves reranking pipelines that score higher and generalize better to new datasets. More broadly, it helps the search better navigate trade-offs between competing objectives, such as quality against cost in reranking. Component-aware feedback provides a practical way to make self-evolving systems more efficient, complementary to improving the search itself.

\bibliography{iclr2027_conference}
\bibliographystyle{iclr2027_conference}

\appendix

\definecolor{pframe}{HTML}{34568B}   
\definecolor{pback}{HTML}{F5F8FD}
\definecolor{phead}{HTML}{1F5FAE}
\definecolor{cframe}{HTML}{1A7F8E}   
\definecolor{cback}{HTML}{F1F8FA}
\lstdefinelanguage{prompt}{
  morecomment=[l][\color{phead}\bfseries]{\#},   
  moredelim=[is][\bfseries]{**}{**}                
}
\lstset{
  basicstyle=\fontencoding{T1}\fontfamily{zi4}\selectfont\scriptsize,  
  columns=flexible, keepspaces=true, breaklines=true, breakindent=1em,
  aboveskip=0pt, belowskip=0pt, upquote=true, showstringspaces=false,
  literate={→}{{$\rightarrow$}}1 {←}{{$\leftarrow$}}1 {Δ}{{$\Delta$}}1 {—}{{---}}1
}
\tcbset{appxbox/.style={breakable, enhanced, listing only, boxrule=0.6pt, arc=3pt,
  left=6pt, right=6pt, top=4pt, bottom=4pt, toptitle=2pt, bottomtitle=2pt,
  fonttitle=\bfseries\small, coltitle=white}}
\newtcblisting[auto counter]{promptbox}[2]{appxbox, label={#2},
  colframe=pframe, colback=pback, colbacktitle=pframe,
  title={Prompt~\thetcbcounter: #1}, listing options={language=prompt}}
\newtcblisting[auto counter]{codebox}[2]{appxbox, label={#2},
colframe=cframe, colback=cback, colbacktitle=cframe,
title={Program~\thetcbcounter: #1},
listing options={language=Python, keywordstyle=\color{cframe}\bfseries,
commentstyle=\color{gray}, stringstyle=\color{orange!70!black}}}

\section{Prompts}
\label{appx:prompts}
The search makes two kinds of LLM calls. Every iteration, a mutation call edits a parent program. When the search stagnates, a strategy call proposes new directions for the mutation calls to implement. Both searches use the same system prompt (Prompt~\ref{prompt:system}) and the same exploration and exploitation blocks (Prompts~\ref{prompt:explore} and~\ref{prompt:exploit}). All examples come from biology runs.

\subsection{System Prompt}
\label{appx:sys-prompt}
The system prompt states the task, the function the program must implement, and the tools the program can call. Only the dataset name changes between datasets.
\begin{promptbox}{System prompt}{prompt:system}
You are an expert tasked with iteratively improving a solution.
Your goal is to maximize the COMBINED SCORE while exploring diverse approaches.
The system maintains a collection of diverse solutions - both high combined score AND diversity are valuable.

The task: given a biology search query, return document ids ranked best-first. Scoring is
nDCG@10 against hidden gold labels. Reading gold labels or importing the evaluator or
dataset is prohibited.

CONTRACT — the evaluator calls exactly this function (keep its name and signature):
    rank(query: str, tools) -> list[str]        # doc_ids, most -> least relevant

TOOLS (passed in as `tools`; do not import anything to get these):
    tools.bm25(query_text: str, k: int) -> list[str]   # BM25 top-k over the corpus; callable repeatedly
    tools.doc(doc_id: str) -> str                      # full document text
    tools.llm(messages: list[dict], max_tokens: int = 1024) -> str   # one chat completion; greedy, token-capped
\end{promptbox}

\subsection{Mutation prompt}
\label{appx:mut-prompt}
Every iteration makes one mutation call. Depending on whether the search is exploring or exploiting, the mutation prompt can include a exploration block (Prompt~\ref{prompt:explore}) or an exploitation block (Prompt~\ref{prompt:exploit}).
\begin{promptbox}{Mutation prompt, exploration block}{prompt:explore}
## PARENT SELECTION CONTEXT
This parent was selected through diversity-driven sampling to explore different regions.

### EXPLORATION GUIDANCE
- Consider alternative algorithmic approaches
- Don't be constrained by the parent's approach
- Look for fundamentally different algorithms or novel techniques
- Balance creativity with correctness

Your goal: Discover new approaches that might outperform current solutions.
\end{promptbox}

\begin{promptbox}{Mutation prompt, exploitation block}{prompt:exploit}
## PARENT SELECTION CONTEXT
This parent was selected from the archive of top-performing programs.

### OPTIMIZATION GUIDANCE
- This solution works well, but meaningful improvements are still possible
- You may refine the existing approach OR introduce better algorithms
- Consider: algorithmic improvements, better data structures, efficient libraries
- Ensure correctness is maintained

Your goal: Improve upon this solution.
\end{promptbox}

Prompts~\ref{prompt:mutation-adaevolve} and~\ref{prompt:mutation-component} are examples of full mutation prompts from AdaEvolve and component-aware feedback.

\begin{promptbox}{AdaEvolve mutation prompt}{prompt:mutation-adaevolve}
# Current Solution Information
- Main Metrics: 
- combined_score: 0.4569

Metrics:
  - ndcg_at_10: 0.4569
  - num_queries: 62
- Focus areas: - Consider simplifying - solution length exceeds 500 characters

# Program Generation History
## Previous Attempts

No previous attempts yet.

## Other Context Solutions
These programs represent diverse approaches and creative solutions that may be relevant to the current task:

### Program 1 (combined_score: 0.3837)
Score breakdown:  - ndcg_at_10: 0.3837  - num_queries: 62

```python
[... 36 lines of program source ...]
```

### Program 2 (combined_score: 0.4389)
Score breakdown:  - ndcg_at_10: 0.4389  - num_queries: 62

```python
[... 66 lines of program source ...]
```

### Program 3 (combined_score: 0.4569)
Score breakdown:  - ndcg_at_10: 0.4569  - num_queries: 62

```python
[... 84 lines of program source ...]
```

### Program 4 (combined_score: 0.4569)
Score breakdown:  - ndcg_at_10: 0.4569  - num_queries: 62

```python
[... 84 lines of program source ...]
```

# Current Solution

## PARENT SELECTION CONTEXT
This parent was selected from the archive of top-performing programs.

### OPTIMIZATION GUIDANCE
- This solution works well, but meaningful improvements are still possible
- You may refine the existing approach OR introduce better algorithms
- Consider: algorithmic improvements, better data structures, efficient libraries
- Ensure correctness is maintained

Your goal: Improve upon this solution.

## Program Information
combined_score: 0.4569
Score breakdown:
  - ndcg_at_10: 0.4569
  - num_queries: 62

```python
[... 84 lines of program source ...]
```

## BREAKTHROUGH IDEA - IMPLEMENT THIS

The search has stagnated globally. You MUST implement this breakthrough idea:

**IDEA:** Explicit Relevance Scoring with LLM

**HOW TO IMPLEMENT:**
Replace the sliding-window listwise reranker with an explicit relevance scoring approach. Retrieve top 50 candidates via BM25 and query expansion, then prompt the LLM to assign a numeric relevance score (1-10) to each document individually. Parse these scores and use the sorted function to order candidates by score descending, returning the top 10. This directly optimizes for nDCG@10 by focusing the LLM's reasoning on absolute relevance rather than relative ordering of many items, which reduces token consumption and eliminates ordering drift in lower-ranked positions. By scoring each doc independently, you avoid context window limitations and ensure the most relevant documents bubble to the top based on explicit evidence.

**TARGET METRIC:** nDCG@10 precision and top-10 recall

**CAUTIONS:** Ensure the prompt strictly requests a structured numeric output to allow reliable parsing with regex. Handle cases where the LLM returns non-numeric values by defaulting to a low score. Keep the candidate pool small (30-50) to stay within token limits while maintaining sufficient recall.

**APPROACH TYPE:** sorted

**CRITICAL:**
- You MUST implement the breakthrough idea
- Ensure the paradigm is actually used in your implementation (not just mentioned in comments)
- Correctness is essential - your implementation must be correct and functional
- Verify output format matches evaluator requirements
- Make purposeful changes that implement the idea
- Test your implementation logic carefully

## PREVIOUS ATTEMPTS ON THIS PARENT
Summary: 0 improved, 0 unchanged, 1 regressed
  1. 0.4569 -> 0.2710 (-0.1859) [REGRESSED]
Avoid repeating approaches that didn't work.

# Task
Suggest improvements to the program that will improve its COMBINED_SCORE.
The system maintains diversity across these dimensions: score, complexity.
Different solutions with similar combined_score but different features are valuable.

You MUST use the exact SEARCH/REPLACE diff format shown below to indicate changes:

<<<<<<< SEARCH
# Original code to find and replace (must match exactly)
=======
# New replacement code
>>>>>>> REPLACE

Example of valid diff format:
<<<<<<< SEARCH
for i in range(m):
    for j in range(p):
        for k in range(n):
            C[i, j] += A[i, k] * B[k, j]
=======
# Reorder loops for better memory access pattern
for i in range(m):
    for k in range(n):
        for j in range(p):
            C[i, j] += A[i, k] * B[k, j]
>>>>>>> REPLACE

**CRITICAL**: You can suggest multiple changes. Each SEARCH section must EXACTLY match code in "# Current Solution" - copy it character-for-character, preserving all whitespace and indentation. Do NOT paraphrase or reformat.
Be thoughtful about your changes and explain your reasoning thoroughly.
Include a concise docstring at the start of functions describing the exact approach taken.

IMPORTANT: If an instruction header of "## IMPORTANT: ..." is given below the "# Current Solution", you MUST follow it. Otherwise, 
focus on targeted improvements of the program. 

- Time limit: Programs should complete execution within 1500 seconds; otherwise, they will timeout.
\end{promptbox}

\begin{promptbox}{Mutation prompt with component-aware feedback}{prompt:mutation-component}
# Current Solution Information
- Main Metrics: 
- combined_score: 0.3942

Metrics:
  - ndcg_at_10: 0.3942
  - num_queries: 62
  - retrieval_recall: 0.4274
  - oracle_ndcg: 0.4813
  - scoring_realization: 0.8190
  - llm_calls_per_query: 18.0000
  - llm_tokens_per_query: 53770.3710
- Focus areas: - Consider simplifying - solution length exceeds 500 characters

# Program Generation History
## What this search has already tried

### Best so far (0.3903)
- made at it8: ~_parse_ranking, ~_rerank_window, ~STRIDE 10->5, ~WINDOW 20->15   combined_score 0.3801 → 0.3903 (Δ +0.0103)
      llm_calls_per_query 9.0000 → 18.0000   llm_tokens_per_query 34707.4032 → 60074.0968   oracle_ndcg 0.4813 → 0.4813   retrieval_recall 0.4274 → 0.4274   scoring_realization 0.7896 → 0.8110
  its parts:
    _parse_ranking:
        [... 52 lines of source ...]
    _rerank_window:
        [... 29 lines of source ...]
    rank:
        [... 16 lines of source ...]

### Other programs in the archive (2)
- program (0.3868)
  made at it9: +_rerank_window, ~_parse_ranking   combined_score 0.2781 → 0.3868 (Δ +0.1087) (+rewired `rank`)
      llm_calls_per_query 1.0000 → 18.0000   llm_tokens_per_query 4213.5968 → 53108.7742   oracle_ndcg 0.3200 → 0.4813   retrieval_recall 0.2737 → 0.4274   scoring_realization 0.8692 → 0.8036
  its parts:
    _parse_ranking:
        [... 19 lines of source ...]
    _rerank_window:
        [... 21 lines of source ...]
    rank:
        [... 13 lines of source ...]
- program (0.3801)
  its parts:
    _parse_ranking:
        [... 11 lines of source ...]
    _rerank_window:
        [... 24 lines of source ...]
    rank:
        [... 16 lines of source ...]

### Changes, and what each one moved  (most recent 10)
[... 7 earlier edits ...]
- it10: ~_parse_ranking, ~_rerank_window, +STRIDE=5, ~WINDOW 25->15   combined_score 0.2943 → 0.3834 (Δ +0.0891) (+rewired `rank`)
      llm_calls_per_query 1.0000 → 18.0000   llm_tokens_per_query 4984.4032 → 60007.6452   oracle_ndcg 0.4813 → 0.4813   retrieval_recall 0.4274 → 0.4274   scoring_realization 0.6114 → 0.7964
    _parse_ranking:
        Parse LLM ranking output into 0-based indices.
        Tries bracket format, numbered lists, comma-separated; fills missing with original order.
    _rerank_window:
        Reorder window_ids by relevance using LLM listwise ranking.
        Uses structured prompt with reasoning instruction. Falls back to input order on failure.
- it11: -STRIDE, ~DEPTH 100->50, ~WINDOW 15->20   combined_score 0.3834 → 0.2847 (Δ -0.0987) (+rewired `rank`)
      llm_calls_per_query 18.0000 → 1.0000   llm_tokens_per_query 60007.6452 → 4470.6613   oracle_ndcg 0.4813 → 0.3996   retrieval_recall 0.4274 → 0.3505   scoring_realization 0.7964 → 0.7124
- it15: ~_parse_ranking, ~_rerank_window   combined_score 0.3868 → 0.3779 (Δ -0.0089) (+rewired `rank`)
      llm_calls_per_query 18.0000 → 18.0000   llm_tokens_per_query 53108.7742 → 55624.3226   oracle_ndcg 0.4813 → 0.4813   retrieval_recall 0.4274 → 0.4274   scoring_realization 0.8036 → 0.7852
    _parse_ranking:
        Parse LLM ranking output into a valid permutation of 0..k-1.
        Prioritizes explicit bracket format [a] > [b] > [c], then falls back to
        numbered lists. Always returns a valid permutation by filling gaps with
        original order.
    _rerank_window:
        Reorder window_ids by relevance using LLM listwise ranking.
        Uses a system message for better instruction following and a structured
        prompt with explicit format constraints. Falls back to input order on failure.

# Current Solution

## PARENT SELECTION CONTEXT
This parent was selected from the archive of top-performing programs.

### OPTIMIZATION GUIDANCE
- This solution works well, but meaningful improvements are still possible
- You may refine the existing approach OR introduce better algorithms
- Consider: algorithmic improvements, better data structures, efficient libraries
- Ensure correctness is maintained

Your goal: Improve upon this solution.

## Program Information
combined_score: 0.3942
Score breakdown:
  - ndcg_at_10: 0.3942
  - num_queries: 62
  - retrieval_recall: 0.4274
  - oracle_ndcg: 0.4813
  - scoring_realization: 0.8190
  - llm_calls_per_query: 18.0000
  - llm_tokens_per_query: 53770.3710

```python
[... 58 lines of program source ...]
```

## BREAKTHROUGH IDEA - IMPLEMENT THIS

The search has stagnated globally. You MUST implement this breakthrough idea:

**IDEA:** Reciprocal Rank Fusion (RRF) for Multi-Query Retrieval

**HOW TO IMPLEMENT:**
Biology queries often miss relevant documents due to terminology gaps. Use the LLM to generate 3-5 semantically related queries. Run tools.bm25 for the original and each expanded query. Merge all candidate pools using Reciprocal Rank Fusion (RRF), scoring each document as the sum of 1 / (k + rank) across all queries (with k=60). This single fusion step captures lexical and semantic diversity in one pass, directly boosting retrieval recall and oracle NDCG without complex pipelines.

**TARGET METRIC:** retrieval_recall, oracle_ndcg

**CAUTIONS:** Keep expanded queries concise to stay within token limits. Use a fixed RRF constant. Ensure strict deduplication while preserving rank positions. Handle cases where BM25 returns fewer than k results gracefully.

**APPROACH TYPE:** custom_rrf_fusion

**CRITICAL:**
- You MUST implement the breakthrough idea
- Ensure the paradigm is actually used in your implementation (not just mentioned in comments)
- Correctness is essential - your implementation must be correct and functional
- Verify output format matches evaluator requirements
- Make purposeful changes that implement the idea
- Test your implementation logic carefully

# Task
[... identical to the Task section of Prompt 4 ...]
\end{promptbox}

\subsection{Strategy prompt}
\label{appx:strat-prompt}
When the search stagnates, a strategy call proposes three new strategies. Later mutation calls implement them on the best program. AdaEvolve's strategy prompt shows the best program and a list of strategies already tried. Component-aware feedback adds the seed as a reference point.

\begin{promptbox}{AdaEvolve strategy prompt}{prompt:strategy-adaevolve}
## Problem Objective

You are an expert tasked with iteratively improving a solution.
Your goal is to maximize the COMBINED SCORE while exploring diverse approaches.
The system maintains a collection of diverse solutions - both high combined score AND diversity are valuable.

The task: given a biology search query, return document ids ranked best-first. Scoring is
nDCG@10 against hidden gold labels. Reading gold labels or importing the evaluator or
dataset is prohibited.

CONTRACT — the evaluator calls exactly this function (keep its name and signature):
    rank(query: str, tools) -> list[str]        # doc_ids, most -> least relevant

TOOLS (passed in as `tools`; do not import anything to get these):
    tools.bm25(query_text: str, k: int) -> list[str]   # BM25 top-k over the corpus; callable repeatedly
    tools.doc(doc_id: str) -> str                      # full document text
    tools.llm(messages: list[dict], max_tokens: int = 1024) -> str   # one chat completion; greedy, token-capped


## Optimization Targets

Optimize the primary scalar score defined by the evaluator.

## Evaluator Code (shows how solutions are scored)

```python
[... evaluator source ...]
```

## Current Best Program (score: 0.483734)

```python
[... 90 lines of program source ...]
```

**CRITICAL:** Analyze the current program first. What algorithm does it use?
What are its strengths and weaknesses? How can you improve upon it?

**CRITICAL: ANALYZE THE CURRENT PROGRAM FIRST**
Before suggesting new ideas, carefully analyze the Current Program above:
- What algorithm/approach does it use? (This is what's WORKING - score 0.483734)
- What are its strengths? (Why does it achieve this score?)
- What are its weaknesses? (What limits further improvement?)
- How can you improve it? (How to beat it?)

**IMPORTANT:** The program above is the CURRENT program that needs to be improved. Start by understanding what works, then suggest breakthrough ideas that build on or improve it.

## Analysis Framework - Complete Before Generating Ideas

**STEP 0: Understand the TASK (MOST IMPORTANT - DO THIS FIRST)**
- What is the problem asking you to do?
- What is the goal or objective? (maximize, minimize, optimize)
- What are the inputs and outputs?
- What needs to be improved? (variables/decisions that affect the goal)
- What constraints exist?

**STEP 1: Analyze the Evaluator Code**
- How are solutions scored?
- What metrics are computed?
- What causes failures or penalties?

**STEP 2: Identify Metrics**
- What is the primary metric or Pareto objective set?
- How is it calculated?
- What secondary metrics exist?
- If variance/std is penalized, the program needs consistency across scenarios

**STEP 3: Identify Constraints**
- What conditions must be satisfied?
- What validation happens?
- What causes score penalties?

**STEP 4: Identify Problem Structure**
- Is processing sequential or global?
- Are decision variables discrete or continuous?
- What dependencies exist between decisions?
- **CRITICAL:** What data does your program receive vs what the evaluator uses?
- **CRITICAL:** How are metrics computed across components? Independently then aggregated, or jointly?

**STEP 5: Determine Appropriate Approach**
- Match approach to problem structure
- Consider what has worked vs failed before
- Identify promising library/technique combinations

**STEP 6: Identify Improvement Opportunities**
- What would increase each metric?
- What would satisfy constraints better?
- What fundamentally different approaches could work?

Current best score is 0.483734. Your ideas must improve the configured optimization targets and, in multiobjective mode, explicitly reason about objective trade-offs.

## Previously Tried Ideas - CHECK THIS FIRST

**CRITICAL:** Review what was already tried. Do NOT suggest ideas that use
the same libraries, functions, or approaches as FAILED attempts.

- FAILED: sorted - Explicit Relevance Scoring with LLM (improvement: +0.0000)
- FAILED: set.union - Multi-Query Entity Expansion for Recall (improvement: +0.0000)
- FAILED: itertools.combinations - Pairwise Knockout Tournament for Top-10 Selection (improvement: +0.0000)
- SUCCESS: scipy.ndimage.generic_filter - Multi-stage retrieval funnel with explicit top-10 optimization (improvement: +0.0268)
- SUCCESS: numpy.fft.fft - Query understanding and adaptive retrieval using LLM-generated search terms (improvement: +0.0268)
- SUCCESS: scipy.cluster.hierarchy.linkage - Iterative retrieval with gap analysis and complementary query generation (improvement: +0.0268)

**STRICT PROHIBITION:** Do NOT keep suggesting approaches that have already failed.
If an approach failed, understand WHY before suggesting similar techniques.
Prioritize approaches that are fundamentally different from failed attempts.

**Learning from Failures - Understand Root Causes:**
When a technique fails badly (score decreased significantly), understand WHY before suggesting alternatives:
- **Fundamental mismatch:** Wrong problem type (e.g., continuous optimizer on discrete problem) -> avoid that entire class of approaches
- **Structural mismatch:** Wrong approach for problem structure (e.g., linear proxy for non-linear objective) -> use approaches that match the actual structure
- **Implementation issues:** If the same library failed multiple times or very badly (>10% decrease), it likely indicates a fundamental mismatch - suggest a different class of approaches

## Technique Guidance

**Note:** Standard scientific libraries (scipy, numpy, etc.) are available. PyTorch and TensorFlow are not available.

**For Continuous Optimization with Constraints:**
- scipy.optimize.minimize with constraint handling (SLSQP, trust-constr)
- Multiple initial guesses for global optimization
- Geometric approaches (Voronoi, convex hull)

**For Discrete/Combinatorial Problems:**
- Greedy heuristics with good ordering
- Local search (swaps, moves)
- scipy.optimize.linear_sum_assignment for assignment problems
- scipy.optimize.linprog for linear constraints

**For Graph/Network Problems:**
- NetworkX algorithms (shortest path, min spanning tree, flow)
- Spectral methods (eigenvalue-based ordering)

**For Repair/Reconstruction:**
- Heuristic-based detection and correction
- Structural constraint exploitation
- Averaging/interpolation for consistency

**For Robust Filtering/Noise Reduction:**
- scipy.signal (medfilt, savgol_filter, wiener) for direct filtering
- Use methods that handle outliers better than mean-based (median, percentile)
- Do NOT use scipy.optimize.minimize to tune filter parameters
- Use filtering functions directly, not multi-stage optimization

**General Principles:**
- Prefer single-function library calls over multi-stage pipelines
- Match algorithm to problem structure
- Simple approaches with good heuristics often beat complex methods

## ANTI-PATTERNS - Critical rules about what NOT to do

1. **Do NOT use multi-stage optimization**: Do NOT call one function then optimize its output. Deterministic setup code followed by a single optimization call is allowed.

2. **Do NOT use scipy.optimize.minimize for hyperparameter tuning**: Use minimize to solve the problem directly, NOT to tune parameters for another function.

3. **Do NOT use scipy.optimize.minimize for discrete problems**: Continuous optimizers cannot handle discrete constraint violations properly.

4. **Each idea MUST be a single-function library call**: Do NOT suggest multi-stage processing (e.g., "call A then call B").

**AVOID:** DEAP, genetic algorithm libraries, domain-specific complex libraries, custom research algorithms, or any library requiring additional `pip install`

**Learning from Success:**
When an approach succeeds, think: what principle made it work? Learn and think of better ideas, don't just add complexity. If breakthrough patterns are known, prioritize approaches that match them.

## DIVERSITY REQUIREMENTS

Before generating ideas, explicitly think:
- Idea 1: [Type A - e.g., algorithmic refinement or library-based approach]
- Idea 2: [Type B - e.g., structural change or processing pattern - DIFFERENT from A]
- Idea 3: [Type C - e.g., different technique or optimization method - DIFFERENT from A and B]

**Verify:** Are these DIFFERENT types? NOT variations of the same approach.

Each idea must:
- Use DIFFERENT libraries/techniques than failed attempts
- Target DIFFERENT metrics/aspects from the evaluator
- Be independently implementable
- Prefer clear implementations (different != more complex)

### Be Specific and Actionable

Not vague: "Try optimization"
Specific: "Use scipy.optimize.minimize with SLSQP method"

- Include exact library names, function names, methods, parameters
- Provide step-by-step implementation guide
- Focus on core logic that implements the idea correctly
- Handle edge cases and avoid errors/warnings
- For optimization: use multiple initializations, appropriate iteration counts and convergence criteria (evaluation timeout: {self.eval_timeout}s)

## Output Format

**IMPORTANT:** Respond with a JSON object containing exactly 3 idea objects under the "ideas" key.
Do not include code patches or diffs — describe strategies in natural language.

Generate 3 breakthrough ideas of DIFFERENT types.

Each idea must be a JSON object with these fields:
- "idea": Clear, direct description with library/technique name
- "description": Detailed implementation guide (5-10 sentences)
- "what_to_optimize": What metrics/areas to focus on
- "cautions": Important implementation details to watch for
- "approach_type": Exact "library.function" format (e.g., "scipy.optimize.minimize")

**Diversity Requirement:** Each idea must use a DIFFERENT approach type.
Do not generate variations of the same technique.

Return ONLY a JSON object in this shape: {"ideas": [ ... ]} with 3 paradigm objects. No other text.

Example:
```json
{
    "ideas": [
        {
            "idea": "Use scipy.optimize.minimize with SLSQP",
            "description": "Apply scipy.optimize.minimize directly to optimize all variables together...",
            "what_to_optimize": "primary evaluator score",
            "cautions": "Ensure constraints are properly formulated, use multiple starting points",
            "approach_type": "scipy.optimize.minimize"
        }
    ]
}
```
\end{promptbox}

\begin{promptbox}{Strategy prompt with component-aware feedback}{prompt:strategy-component}
## Problem Objective

You are an expert tasked with iteratively improving a solution.
Your goal is to maximize the COMBINED SCORE while exploring diverse approaches.
The system maintains a collection of diverse solutions - both high combined score AND diversity are valuable.

The task: given a biology search query, return document ids ranked best-first. Scoring is
nDCG@10 against hidden gold labels. Reading gold labels or importing the evaluator or
dataset is prohibited.

CONTRACT — the evaluator calls exactly this function (keep its name and signature):
    rank(query: str, tools) -> list[str]        # doc_ids, most -> least relevant

TOOLS (passed in as `tools`; do not import anything to get these):
    tools.bm25(query_text: str, k: int) -> list[str]   # BM25 top-k over the corpus; callable repeatedly
    tools.doc(doc_id: str) -> str                      # full document text
    tools.llm(messages: list[dict], max_tokens: int = 1024) -> str   # one chat completion; greedy, token-capped


## Optimization Targets

Optimize the primary scalar score defined by the evaluator.

## Evaluator Code (shows how solutions are scored)

```python
[... 383 lines of program source ...]
```

## The Seed Program, where the search starts (score: 0.380053)

```python
[... 67 lines of program source ...]
```

**NOTE:** This seed is the REFERENCE FRAME, not the program to improve. Every
number in the search history below is stated as a delta from this program.
The program you are improving is the Current Best Program shown next.

## The Current Best Program (score: 0.607106, found at iteration 20)

**This is the program your idea will be applied to.** It is what the seed above became after
the search; read it before proposing anything, because an idea it already implements is a
wasted iteration.

```python
[... 120 lines of program source ...]
```

**CRITICAL: ANALYZE THE CURRENT PROGRAM FIRST**
Before suggesting new ideas, carefully analyze the Current Program above:
- What algorithm/approach does it use? (This is what's WORKING - score 0.607106)
- What are its strengths? (Why does it achieve this score?)
- What are its weaknesses? (What limits further improvement?)
- How can you improve it? (How to beat it?)

**IMPORTANT:** The program above is the CURRENT program that needs to be improved. Start by understanding what works, then suggest breakthrough ideas that build on or improve it.

## Analysis Framework - Complete Before Generating Ideas

**STEP 0: Understand the TASK (MOST IMPORTANT - DO THIS FIRST)**
- What is the problem asking you to do?
- What is the goal or objective? (maximize, minimize, optimize)
- What are the inputs and outputs?
- What needs to be improved? (variables/decisions that affect the goal)
- What constraints exist?

**STEP 1: Analyze the Evaluator Code**
- How are solutions scored?
- What metrics are computed?
- What causes failures or penalties?

**STEP 2: Identify Metrics**
- What is the primary metric or Pareto objective set?
- How is it calculated?
- What secondary metrics exist?
- If variance/std is penalized, the program needs consistency across scenarios

**STEP 3: Identify Constraints**
- What conditions must be satisfied?
- What validation happens?
- What causes score penalties?

**STEP 4: Identify Problem Structure**
- Is processing sequential or global?
- Are decision variables discrete or continuous?
- What dependencies exist between decisions?
- **CRITICAL:** What data does your program receive vs what the evaluator uses?
- **CRITICAL:** How are metrics computed across components? Independently then aggregated, or jointly?

**STEP 5: Determine Appropriate Approach**
- Match approach to problem structure
- Consider what has worked vs failed before
- Identify promising library/technique combinations

**STEP 6: Identify Improvement Opportunities**
- What would increase each metric?
- What would satisfy constraints better?
- What fundamentally different approaches could work?

Current best score is 0.380053. Your ideas must improve the configured optimization targets and, in multiobjective mode, explicitly reason about objective trade-offs.

## What the search has explored, measured from the seed

seed: combined_score 0.3801   llm_calls_per_query 9.0000   llm_tokens_per_query 34707.4032   oracle_ndcg 0.4813   retrieval_recall 0.4274   scoring_realization 0.7896

A named [Strategy] in a lineage is a paradigm; an unnamed one is ordinary
mutation, which is how the search starts. Every figure is against the seed.

CURRENT BEST (combined_score 0.6071 at it20) -- a new idea gets applied to this program:
- seed → 3mut → [Reciprocal Rank Fusion (RRF) for Multi-Query Retrieval](it17) → [Reciprocal Rank Fusion (RRF) for Multi-Query Retrieval](it20)
      combined_score 0.3801 → 0.6071   (Δ +0.2271 from seed)
      llm_calls_per_query 9.0000 → 63.2742   llm_tokens_per_query 34707.4032 → 146786.0806   oracle_ndcg 0.4813 → 0.8036   retrieval_recall 0.4274 → 0.7664   scoring_realization 0.7896 → 0.7555
      net change from seed: +_generate_queries, +_rrf_fuse, ~_parse_ranking, ~_rerank_window, -DEPTH, -STRIDE, -WINDOW (+rewired `rank`)
      _generate_queries:
          Generate 3 semantically related queries using LLM for multi-query retrieval.
          Uses LLM to expand the original query into related formulations that capture
          different terminology, improving retrieval recall across diverse biology concepts.
      _rrf_fuse:
          Fuse multiple BM25 result lists using Reciprocal Rank Fusion.
          Scores each document as sum of 1/(k + rank) across all queries.
          Higher scores indicate better overall ranking across queries.
      _parse_ranking:
          Extract ranking order from LLM output, prioritizing [n] brackets.
      _rerank_window:
          Rerank a window of candidates using LLM listwise comparison.

The 10 most recent lineages, oldest first:
- seed → 3mut → [Reciprocal Rank Fusion (RRF) for Multi-Query Retrieval](it17) → [Reciprocal Rank Fusion (RRF) for Multi-Query Retrieval](it20) → [Tournament-Style Pairwise Elimination for Top-10 Ranking](it31)
      combined_score 0.3801 → 0.3810   (Δ +0.0010 from seed)
      llm_calls_per_query 9.0000 → 45.0000   llm_tokens_per_query 34707.4032 → 24599.2581   oracle_ndcg 0.4813 → 0.5824   retrieval_recall 0.4274 → 0.4986   scoring_realization 0.7896 → 0.6543
      net change from seed: +_generate_queries, +_listwise_rank, +_llm_pairwise, +_pairwise_tournament, +_rrf_fuse, -_rerank_window, ~_parse_ranking, -DEPTH, -STRIDE, -WINDOW (+rewired `rank`)
      _listwise_rank:
          Listwise rank candidates using a single LLM comparison call.
          Takes a small set of candidates (~10) and asks the LLM to order them
          from most to least relevant using bracket notation.
      _llm_pairwise:
          Determine which of two documents is more relevant to the query.
          Uses LLM for direct pairwise comparison. Ties are broken deterministically
          by preferring Document A (which appears earlier in the pre-ranked list,
          serving as a BM25/RRF rank tiebreaker).
      _pairwise_tournament:
          Tournament-style pairwise elimination for top-10 ranking.
          Splits candidates into pairs, uses LLM to determine relevance,
          repeats knockout rounds until <=10 candidates remain, then listwise ranks them.
          This logarithmic approach minimizes LLM calls while maximizing top-10 precision.

- seed → 3mut → [Reciprocal Rank Fusion (RRF) for Multi-Query Retrieval](it17) → [Reciprocal Rank Fusion (RRF) for Multi-Query Retrieval](it20) → [Deterministic Query Expansion with Logarithmic RRF Fusion](it32)
      combined_score 0.3801 → 0.3837   (Δ +0.0036 from seed)
      llm_calls_per_query 9.0000 → 32.5161   llm_tokens_per_query 34707.4032 → 85195.2581   oracle_ndcg 0.4813 → 0.4917   retrieval_recall 0.4274 → 0.4373   scoring_realization 0.7896 → 0.7803
      net change from seed: +_deterministic_expand, +_log_rrf_fuse, ~_parse_ranking, ~_rerank_window, -DEPTH, -STRIDE, -WINDOW (+rewired `rank`)
      _deterministic_expand:
          Expand query deterministically by extracting key terms and handling negations.
          Replaces LLM-based query generation with a lightweight, deterministic strategy
          that improves retrieval recall and oracle NDCG without LLM overhead.
      _log_rrf_fuse:
          Fuse multiple BM25 result lists using logarithmic RRF with original query weighting.
          Heavily weights documents from the original query while boosting expanded matches.
          Penalizes documents appearing only in expanded queries to maintain precision.

- seed → 3mut → [Reciprocal Rank Fusion (RRF) for Multi-Query Retrieval](it17) → [Reciprocal Rank Fusion (RRF) for Multi-Query Retrieval](it20) → [Batched Pointwise LLM Relevance Scoring with Structured JSON](it33)
      combined_score 0.3801 → 0.3911   (Δ +0.0110 from seed)
      llm_calls_per_query 9.0000 → 2.0000   llm_tokens_per_query 34707.4032 → 8821.9194   oracle_ndcg 0.4813 → 0.6531   retrieval_recall 0.4274 → 0.6003   scoring_realization 0.7896 → 0.5988
      net change from seed: +_batch_score_candidates, +_generate_queries, +_rrf_fuse, -_parse_ranking, -_rerank_window, -DEPTH, -STRIDE, -WINDOW (+rewired `rank`)
      _batch_score_candidates:
          Score candidates by relevance using a single batched LLM call with JSON output.
          Evaluates each candidate on a 0-100 relevance scale using biology-specific
          criteria (accuracy, direct relevance, completeness, specificity). Returns
          candidates sorted by score descending. Single LLM call eliminates error
          propagation from window overlaps and reduces calls from ~60 to 1.

[... 7 more lineages ...]

## Technique Guidance

**Note:** Standard scientific libraries (scipy, numpy, etc.) are available. PyTorch and TensorFlow are not available.

**For Continuous Optimization with Constraints:**
- scipy.optimize.minimize with constraint handling (SLSQP, trust-constr)
- Multiple initial guesses for global optimization
- Geometric approaches (Voronoi, convex hull)

**For Discrete/Combinatorial Problems:**
- Greedy heuristics with good ordering
- Local search (swaps, moves)
- scipy.optimize.linear_sum_assignment for assignment problems
- scipy.optimize.linprog for linear constraints

**For Graph/Network Problems:**
- NetworkX algorithms (shortest path, min spanning tree, flow)
- Spectral methods (eigenvalue-based ordering)

**For Repair/Reconstruction:**
- Heuristic-based detection and correction
- Structural constraint exploitation
- Averaging/interpolation for consistency

**For Robust Filtering/Noise Reduction:**
- scipy.signal (medfilt, savgol_filter, wiener) for direct filtering
- Use methods that handle outliers better than mean-based (median, percentile)
- Do NOT use scipy.optimize.minimize to tune filter parameters
- Use filtering functions directly, not multi-stage optimization

**General Principles:**
- Prefer single-function library calls over multi-stage pipelines
- Match algorithm to problem structure
- Simple approaches with good heuristics often beat complex methods

## ANTI-PATTERNS - Critical rules about what NOT to do

1. **Do NOT use multi-stage optimization**: Do NOT call one function then optimize its output. Deterministic setup code followed by a single optimization call is allowed.

2. **Do NOT use scipy.optimize.minimize for hyperparameter tuning**: Use minimize to solve the problem directly, NOT to tune parameters for another function.

3. **Do NOT use scipy.optimize.minimize for discrete problems**: Continuous optimizers cannot handle discrete constraint violations properly.

4. **Each idea MUST be a single-function library call**: Do NOT suggest multi-stage processing (e.g., "call A then call B").

**AVOID:** DEAP, genetic algorithm libraries, domain-specific complex libraries, custom research algorithms, or any library requiring additional `pip install`

**Learning from Success:**
When an approach succeeds, think: what principle made it work? Learn and think of better ideas, don't just add complexity. If breakthrough patterns are known, prioritize approaches that match them.

## DIVERSITY REQUIREMENTS

Before generating ideas, explicitly think:
- Idea 1: [Type A - e.g., algorithmic refinement or library-based approach]
- Idea 2: [Type B - e.g., structural change or processing pattern - DIFFERENT from A]
- Idea 3: [Type C - e.g., different technique or optimization method - DIFFERENT from A and B]

**Verify:** Are these DIFFERENT types? NOT variations of the same approach.

Each idea must:
- Use DIFFERENT libraries/techniques than failed attempts
- Target DIFFERENT metrics/aspects from the evaluator
- Be independently implementable
- Prefer clear implementations (different != more complex)

### Be Specific and Actionable

Not vague: "Try optimization"
Specific: "Use scipy.optimize.minimize with SLSQP method"

- Include exact library names, function names, methods, parameters
- Provide step-by-step implementation guide
- Focus on core logic that implements the idea correctly
- Handle edge cases and avoid errors/warnings
- For optimization: use multiple initializations, appropriate iteration counts and convergence criteria (evaluation timeout: {self.eval_timeout}s)

## Output Format

**IMPORTANT:** Respond with a JSON object containing exactly 3 idea objects under the "ideas" key.
Do not include code patches or diffs — describe strategies in natural language.

Generate 3 breakthrough ideas of DIFFERENT types.

Each idea must be a JSON object with these fields:
- "idea": Clear, direct description with library/technique name
- "description": Detailed implementation guide (5-10 sentences)
- "what_to_optimize": What metrics/areas to focus on
- "cautions": Important implementation details to watch for
- "approach_type": Exact "library.function" format (e.g., "scipy.optimize.minimize")

**Diversity Requirement:** Each idea must use a DIFFERENT approach type.
Do not generate variations of the same technique.

Return ONLY a JSON object in this shape: {"ideas": [ ... ]} with 3 paradigm objects. No other text.

Example:
```json
{
    "ideas": [
        {
            "idea": "Use scipy.optimize.minimize with SLSQP",
            "description": "Apply scipy.optimize.minimize directly to optimize all variables together...",
            "what_to_optimize": "primary evaluator score",
            "cautions": "Ensure constraints are properly formulated, use multiple starting points",
            "approach_type": "scipy.optimize.minimize"
        }
    ]
}
```
\end{promptbox}

\section{Programs}
\label{appx:programs}

\subsection{Seed program}
\label{appx:seed-prog}
Both searches initialize every run with this program. It retrieves the top 100 BM25 candidates and reranks them with a sliding-window listwise reranker~\citep{rankgpt}, using windows of 20 with a stride of 10.

\begin{codebox}{Seed program}{prog:seed}
"""Initial BRIGHT reranker — BM25 top-100 + RankGPT sliding-window baseline."""

import re

# Tunables — evolution may change these or remove them entirely.
DEPTH = 100     # BM25 candidates to rerank
WINDOW = 20     # sliding-window size
STRIDE = 10     # sliding-window step


def _parse_ranking(text, k):
    """Parse '[3] > [1] > ...' into 0-based indices: deduped, restricted to 0..k-1, with
    any omitted identifiers appended in original order (always a full permutation)."""
    order, seen = [], set()
    for m in re.findall(r"\[(\d+)\]", text or ""):
        idx = int(m) - 1
        if 0 <= idx < k and idx not in seen:
            seen.add(idx)
            order.append(idx)
    order.extend(i for i in range(k) if i not in seen)
    return order


def _rerank_window(query, window_ids, tools):
    """One listwise LLM call: reorder window_ids by relevance. Falls back to input order
    on any failure, so the window is always a valid permutation of its inputs."""
    passages = "\n\n".join(
        f"[{i + 1}] {tools.doc(cid)}" for i, cid in enumerate(window_ids)
    )
    system = "Rank the passages."
    user = (
        f"Query:\n{query}\n\n"
        f"Candidate passages:\n{passages}\n\n"
        f"Rank ALL {len(window_ids)} passages from most to least relevant to the query. "
        f"Respond ONLY with the ranking as identifiers in the format "
        f"[a] > [b] > [c] > ... using every identifier from 1 to {len(window_ids)} exactly once."
    )
    try:
        text = tools.llm(
            [{"role": "system", "content": system},
             {"role": "user", "content": user}],
            max_tokens=1024,
        )
        order = _parse_ranking(text, len(window_ids))
    except Exception:
        order = list(range(len(window_ids)))
    return [window_ids[i] for i in order]


def rank(query, tools):
    """Return doc_ids ranked most -> least relevant for `query`."""
    # First stage: BM25 candidate pool.
    ranked = tools.bm25(query, DEPTH)

    # Second stage: RankGPT listwise sliding window, bottom -> top so relevant docs
    # bubble upward through the overlap.
    window = min(WINDOW, len(ranked))
    end = len(ranked)
    while window and end > 0:
        start = max(0, end - window)
        ranked[start:end] = _rerank_window(query, ranked[start:end], tools)
        if start == 0:
            break
        end -= STRIDE
    return ranked
\end{codebox}

\subsection{Evolved programs}
\label{appx:evo-prog}
For each objective, we show one of the final programs from biology with the highest train score. The docstrings and comments are written and maintained by the mutation model.

\paragraph{Quality-only objective.} Query expansion with reciprocal rank fusion and a denser sliding window, reaching $0.633$ nDCG@10 on test at $151$k tokens per query.

\begin{codebox}{Evolved program, quality-only objective}{prog:quality}
"""BM25 retrieval with multi-query RRF fusion + sliding window LLM reranker."""

import re


def _generate_queries(query, tools):
    """Generate 3 semantically related queries using LLM for multi-query retrieval.
    
    Uses LLM to expand the original query into related formulations that capture
    different terminology, improving retrieval recall across diverse biology concepts.
    """
    prompt = (
        f"Original query: {query}\n\n"
        f"Generate exactly 3 related search queries for biology research.\n"
        f"Keep each query concise (3-8 words). Output one per line.\n"
        f"Format: 1. query1\n2. query2\n3. query3\n"
        f"Do not include any other text."
    )
    try:
        text = tools.llm(
            [{"role": "system", "content": "You are a helpful assistant."},
             {"role": "user", "content": prompt}],
            max_tokens=256
        )
        # Extract numbered queries
        lines = text.strip().split('\n')
        expanded = []
        for line in lines:
            line = line.strip()
            # Remove numbering
            m = re.match(r"\d+[\.\)]\s*(.+)", line)
            if m:
                expanded.append(m.group(1).strip())
            elif line and not line.startswith('#'):
                expanded.append(line)
        # Keep only first 3
        return [query] + expanded[:3]
    except Exception:
        return [query]


def _rrf_fuse(all_results, k=60):
    """Fuse multiple BM25 result lists using Reciprocal Rank Fusion.
    
    Scores each document as sum of 1/(k + rank) across all queries.
    Higher scores indicate better overall ranking across queries.
    """
    scores = {}
    for rank_list in all_results:
        for rank, doc_id in enumerate(rank_list, 1):
            scores[doc_id] = scores.get(doc_id, 0) + 1.0 / (k + rank)
    # Sort by score descending
    fused = sorted(scores.keys(), key=lambda x: -scores[x])
    return fused


def _parse_ranking(text, k):
    """Extract ranking order from LLM output, prioritizing [n] brackets."""
    order, seen = [], set()
    for m in re.findall(r"\[(\d+)\]", text or ""):
        idx = int(m) - 1
        if 0 <= idx < k and idx not in seen:
            seen.add(idx)
            order.append(idx)
    # Fill missing with original order to guarantee a valid permutation
    order.extend(i for i in range(k) if i not in seen)
    return order[:k]


def _rerank_window(query, window_ids, tools):
    """Rerank a window of candidates using LLM listwise comparison."""
    if not window_ids:
        return []
    passages = "\n\n".join(f"[{i+1}] {tools.doc(cid)}" for i, cid in enumerate(window_ids))
    prompt = (
        f"Query: {query}\n\n"
        f"Candidates:\n{passages}\n\n"
        f"Rank these {len(window_ids)} passages from most to least relevant.\n\n"
        f"Output ONLY the final ranking in this exact format: [a] > [b] > [c]\n"
        f"Use every identifier from 1 to {len(window_ids)} exactly once.\n"
        f"Do not include any other text."
    )
    try:
        text = tools.llm(
            [{"role": "system", "content": "You are a helpful assistant that ranks documents."},
             {"role": "user", "content": prompt}],
            max_tokens=1024
        )
        order = _parse_ranking(text, len(window_ids))
    except Exception:
        order = list(range(len(window_ids)))
    return [window_ids[i] for i in order]


def rank(query, tools):
    """Return doc_ids ranked most -> least relevant for `query`."""
    # Multi-query retrieval with RRF fusion
    queries = _generate_queries(query, tools)
    all_results = []
    for q in queries:
        results = tools.bm25(q, 100)
        all_results.append(results)
    
    # Fuse using RRF to boost recall and oracle NDCG
    ranked = _rrf_fuse(all_results, k=60)
    
    if not ranked:
        return []
    
    # Sliding window LLM reranker
    window, stride, end = 15, 5, len(ranked)
    while end > 0:
        start = max(0, end - window)
        ranked[start:end] = _rerank_window(query, ranked[start:end], tools)
        if start == 0:
            break
        end -= stride
    return ranked
\end{codebox}

\paragraph{Cost-aware objective.} This program replaces the sliding window with a single listwise LLM call over the top 100 candidates. Before ranking, it asks the LLM which concepts the top ten BM25 results are missing and retrieves those concepts. It reaches $0.569$ nDCG@10 on test with $2$ LLM calls and $17.5$k tokens per query, about a ninth of the tokens of the quality-only program.

\begin{codebox}{Evolved program, cost-aware objective}{prog:cost}
"""Single-call LLM reranker — BM25 top-100 with comprehensive ranking in one LLM call."""

import re


def _parse_ranking(text, k):
    """Parse '[3] > [1] > ...' into 0-based indices: deduped, restricted to 0..k-1, with
    any omitted identifiers appended in original order (always a full permutation)."""
    order, seen = [], set()
    for m in re.findall(r"\[(\d+)\]", text or ""):
        idx = int(m) - 1
        if 0 <= idx < k and idx not in seen:
            seen.add(idx)
            order.append(idx)
    order.extend(i for i in range(k) if i not in seen)
    return order


def rank(query, tools):
    """Coverage-Greedy Candidate Expansion with Conceptual Gap Analysis."""
    # 1. Initial BM25 retrieval
    initial = tools.bm25(query, 50)
    if not initial:
        return []
        
    # 2. Conceptual Gap Analysis
    # Preview top 10 candidates to identify missing biological concepts
    preview = "\n".join(f"[{i+1}] {tools.doc(d)}" for i, d in enumerate(initial[:10]))
    gap_prompt = (
        f"Query: {query}\n\n"
        f"Top retrieved docs cover:\n{preview}\n\n"
        f"Identify up to 3 important biological concepts missing from these results. "
        f"Return them as a comma-separated list of short search terms."
    )
    try:
        gap_resp = tools.llm([{"role": "user", "content": gap_prompt}], max_tokens=128)
        gap_terms = [t.strip().strip('"\'') for t in gap_resp.split(",") if t.strip()]
    except Exception:
        gap_terms = []
        
    # 3. Targeted retrieval for gaps
    gap_hits = []
    for term in gap_terms[:3]:
        gap_hits.extend(tools.bm25(term, 15))
        
    # 4. RRF Fusion
    rrf = {}
    for i, d in enumerate(initial):
        rrf[d] = rrf.get(d, 0) + 1.0 / (i + 60)
    for d in gap_hits:
        rrf[d] = rrf.get(d, 0) + 1.0 / 60.0
        
    fused = sorted(rrf.items(), key=lambda x: x[1], reverse=True)
    candidates = [d for d, _ in fused[:100]]
    
    if not candidates:
        return []
        
    # 5. LLM Comprehensive Ranking
    passages = "\n".join(f"[{i+1}] {tools.doc(d)}" for i, d in enumerate(candidates))
    prompt = (
        f"Query: {query}\n\n"
        f"Rank ALL {len(candidates)} passages from most to least relevant.\n"
        f"Passages:\n{passages}\n\n"
        f"Respond ONLY with the ranking in the format [1] > [2] > [3] ... "
        f"using every identifier from 1 to {len(candidates)} exactly once."
    )
    
    try:
        text = tools.llm(
            [{"role": "system", "content": "You are a relevance ranking assistant."},
             {"role": "user", "content": prompt}],
            max_tokens=1024,
        )
        order = _parse_ranking(text, len(candidates))
    except Exception:
        order = list(range(len(candidates)))
        
    return [candidates[i] for i in order]
\end{codebox}

\end{document}